# Causal Discovery and Causal Learning for Fire Resistance Evaluation: Incorporating Domain Knowledge

M.Z. Naser[1], Aybike Özyüksel Çiftçioğl[2]

[1]School of Civil & Environmental Engineering and Earth Sciences (SCEEES), Clemson University, USA
[1]Artificial Intelligence Research Institute for Science and Engineering (AIRISE), Clemson University, USA
E-mail: mznaser@clemson.edu, Website: www.mznaser.com

[2]Department of Civil Engineering, Manisa Celal Bayar University, Turkey, E-mail: aybike.ozyuksel@cbu.edu.tr

**Abstract**
Experiments remain the gold standard to establish an understanding of fire-related phenomena. A primary goal in designing tests is to uncover the data generating process (i.e., the how and why the observations we see come to be); or simply what causes such observations. Uncovering such a process not only advances our knowledge but also provides us with the capability to be able to predict phenomena accurately. This paper presents an approach that leverages causal discovery and causal inference to evaluate the fire resistance of structural members. In this approach, causal discovery algorithms are adopted to uncover the causal structure between key variables pertaining to the fire resistance of reinforced concrete (RC) columns. Then, companion inference algorithms are applied to infer (estimate) the influence of each variable on the fire resistance given a specific intervention. Finally, this study ends by contrasting the algorithmic causal discovery with that obtained from domain knowledge and traditional machine learning. Our findings clearly show the potential and merit of adopting causality into our domain.

<u>*Keywords*</u>: Causal discovery; Causal inference; Structural fire engineering; RC columns; Machine learning.

**Introduction**

In our pursuit of discovering knowledge, we seek to identify, or possibly retrieve, the underlying data generating process (DGP) responsible for producing the observations we hope to understand [1]. As such, we devise experiments. Such experiments are designed to test and explore hypotheses. A hypothesis targets a specific direction to uncover the DGP behind a given phenomenon (i.e., the cause(s) leading to the so-called effect). In other words, our experiments are fueled by hypotheses that examine how a set of events/states may produce other events/states. More generally, we look to identify the causal path (i.e., cause(s) → effect), for, without the cause(s), the event would not have been generated[1] [2].

Say that we uncover the true DGP behind fire testing, then it is plausible that this knowledge, once validated, will enable us to focus on other burning questions within our domain. While that is an ambitious goal, uncovering the true DGP is equally important in the short run as it could reduce our reliance on expensive and complex fire tests. For example, instead of testing a series of specimens, one may opt to utilize the identified DGP to estimate the outcome of a particular testing campaign and perhaps test a couple of specimens (vs. all specimens) to cross-check the estimates obtained via the DGP.

Unlike the concepts of the digital twin or finite element (FE) modeling, a causal method is an attractive one-stop approach that may not require continuous simulations (which entails monitoring, collecting data, storage of data, and, most importantly, the building and re-building of case-specific models). As such, it is the opinion of the authors of this work that pursuing this research direction is of high merit.

---

[1] The discussion on causality can continue to the intersection of philosophy, epistemology, and ontology. A broader look into causality can be found herein [43,64,65].

Unfortunately, a search using the *Dimensions* academic database [3] pertaining to the terms "causality", "causal discovery", and "causal inference" when paired with "structural fire engineering" (individually or collectively) returns a minute amount of published work [4]. On a more positive note, the amount of work with a causal theme is much larger when related to social sciences and/or ecological sciences within the fire domain. This observation and companion informetric analysis restraint our literature review discussion on former work while, at the same time, establishes the need for this current study.

In lieu of the above, we opt to highlight some of the key factors and recent efforts that tackled the problem of fire resistance of RC columns. From this perspective, exposure to elevated temperatures has been noted to degrade the properties of construction materials via a series of physio-chemical reactions [5,6]. This often leads to losses in the strength and elastic modulus properties [6]. The loss of strength is primarily a function of the concrete type, among other factors such as heating rate, mixture proportions, etc. [7]. This loss in strength and stiffness properties implies that fire-exposed RC columns are bound to undergo some level of capacity loss under fire conditions.

Notably, RC columns made from normal strength concrete (NSC) display admirable performance under fire conditions – especially when compared to columns made from higher grade concrete materials [8]. In contrast to NSC, high strength concrete (HSC) and ultra high-performance concrete (UHPC) have a much denser structure and low water/cement ratio. Despite the above, surprisingly, the temperature-induced degradation in HSC and UHPC has been shown to occur at a rapid pace [9,10]. Thus, although HSC and UHPC may attain high strength (2-4 times that of NSC), the same strength does not correlate to improved fire resistance. In fact, the correlation between compressive strength and fire resistance of more than 130 RC columns shown in Fig. 1 is a weak positive correlation of 0.21. It is interesting to point out that a column with low concrete strength does not seem to guarantee to achieve high fire resistance.

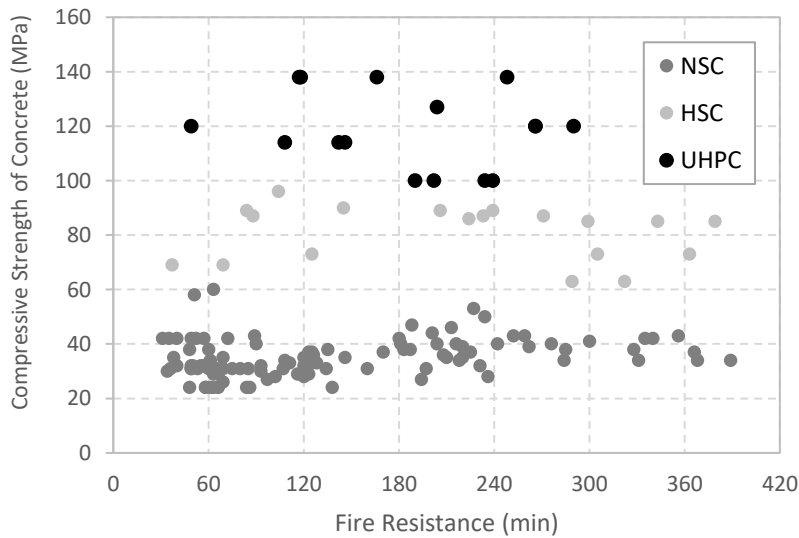

Fig. 1 Examination of compressive strength and fire resistance of fire-tested RC columns

Other factors that are also linked to the fire resistance of RC columns include the geometric features, the level of applied loading, boundary conditions, and fire scenario (heating rate, heating duration, maximum temperature, cooling duration, etc.) [11–13]. The interaction of such factors has been heavily examined in the open literature experimentally [14,15], numerically [16,17], and theoretically [18,19]. Other notable studies worthy of mentioning include those that explored the influence of unsymmetric heating/loading [20–23], unique geometrical features [11], design/natural fire conditions [24], reinforcement configuration [25,26], as well as residual response post fire [27–29].

Practically, the fire resistance of RC columns can be predicted and calculated via a number of approaches. These approaches range from traditional aids (i.e., via charts, tables) as documented in fire buildings and standards [30,31], to hand-calculation based methods derived by researchers [27,32–34], to alternative methods such as finite element simulations [16,26], and/or machine learning (ML) (blackbox algorithms [35–38], or explainable models [39]). The aforenoted methods deliver predictions on the expected fire resistance of RC columns *given* a set of variables. Interestingly, these methods do not often agree if applied to a particular and/or a set of columns – possibly due to differences in derivation, principles, assumptions, and fundamentals [40–42]. The same also presents an opportunity to re-visit the classical phenomenon of fire resistance of RC columns. Indeed, this is the second motivation behind this work.

This paper presents a casual approach to discovering and inferring the causal mechanism responsible for the data generating process of the fire resistance of RC columns. First, casual discovery learning is carried out to uncover the causal structure responsible for tying the variables involved in the fire resistance of RC columns. Four causal structures were identified herein through algorithmic search and incorporation of domain knowledge. Then, inference algorithms are used to estimate the influence of interventions upon the noted variables in each of the identified four structures. For completion, a comparison is drawn to examine the newly discovered knowledge against domain knowledge and traditional machine learning.

**Development of the Proposed Causal Approach**

Regression can be comfortably used to *predict* an outcome of interest, $Y$, given a set of regressors. A prediction in this manner does not imply nor indicate that the regressors, $X$'s, are causes of $Y$ – thu such assignment may arrive from domain knowledge. Assigning such regressors is hardly associated with checking of confounders or common causes of the $X$'s either. On the other hand, a causal analysis strives to establish if a set of variables causes $Y$. A look into Fig. 2 showcases a visual depiction of how regression differs from causation.

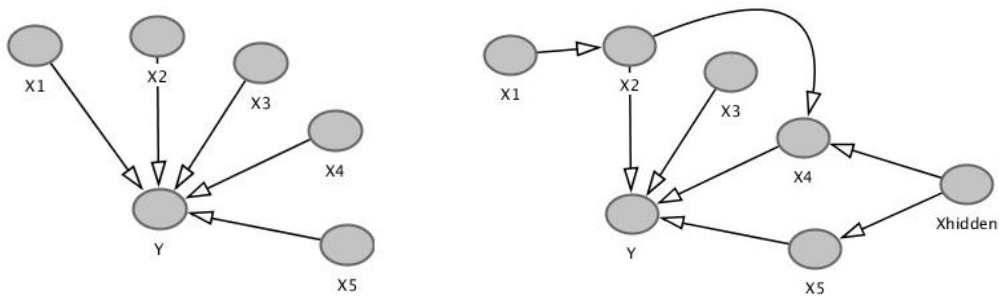

Fig. 2 Regression vs. causation

The proposed causal approach comprises three primary steps. In the first step, causal discovery is adopted to uncover the underlying structure pertaining to the causal ties between a select of variables. Those variables can be identified from domain knowledge, experts' opinions, and/or numerically (i.e., algorithmically). The causal links are established by satisfying causal principles. These principles include the Markov causal assumption, the causal faithfulness assumption, and the causal sufficiency assumption. These assumptions are described herein, and full details can be found elsewhere [43–45].

The first assumption states that a given variable is independent of all other variables (except its *own* effects) conditional on its direct causes. This Markovian assumption is accompanied by the *d-separation* criterion [43], which entails whether a variable is independent of another given a third by associating the notion of independence with the separation of variables in a causal graph. The *casual faithfulness* assumption states that any population produced by a causal graph has the independence relations obtained by applying the d-separation criterion. The c*ausal sufficiency* assumption refers to the absence of hidden or latent parameters that we do not know nor are aware of. For completion, a causal graph is a directed acyclic graph (DAG) that satisfies the causal assumptions – see Fig. 3[2].

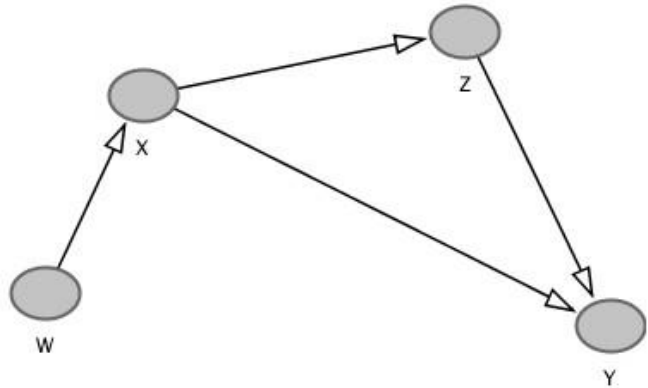

Fig. 3 A sample of a DAG where *W* is assumed to cause *X*, *X* is assumed to cause *Z* and *Y*, and *Z* is also assumed to cause *Y* [Note: *Y* is conditionally independent of *W* given *X,* and *Z* is conditionally independent of *W* given *X*]

In the second step, and once the causal structure (say a DAG) is identified, causal inference algorithms are applied to infer how would the output (i.e., fire resistance of RC columns) change by intervening (e.g., changing) the magnitude or degree of a governing variable. An intervention equates to *setting X = x* as opposed to *observing X = x*. The former relates to causation (what is the fire resistance of a RC column if its width is *increased* to 300 mm?) and the latter to prediction (what is the fire resistance of a RC column *given* it has a width of 300 mm?).

Finally, in the last step, findings from the causal analysis can be compared against that of a companion analysis (say, from traditional machine learning, statistical analysis, domain knowledge, etc.). An advantage to comparing the causal analysis to existing theory may come in handy to further verify the correctness of an adopted theory or in checking if the causal analysis mirrors that from domain knowledge. Figure 4 demonstrates the proposed approach in more detail.

---

[2] The reader is highly encouraged to visit the following sources on causality, do-calculus, and others [66,67].

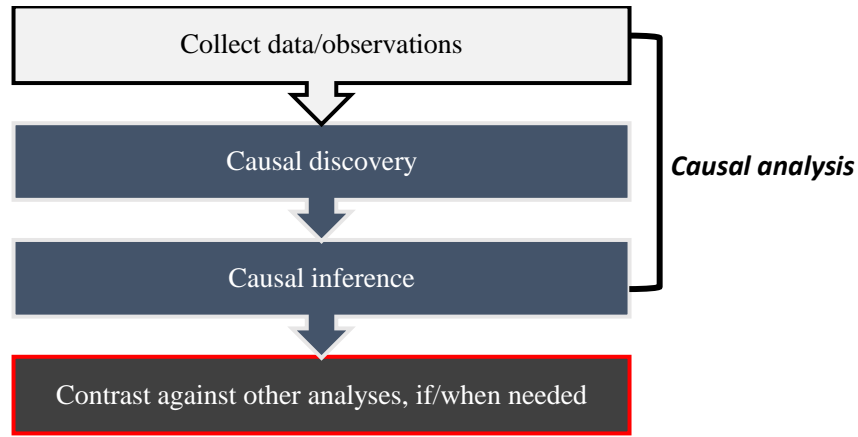

Fig. 4 Flowchart of the proposed approach

**Description of Causal Algorithms and Models**
This section describes the two main causal algorithms and corresponding models showcased herein (CausalNex and DoWhy), and a full description of these tools can be found in their respective references [46,47]. To maintain coherence and allow the replication of our analysis, we adopted both algorithms in their default settings. In a nutshell, we adopted CausalNex to uncover the DGP and then incorporated such DGP into DoWhy to infer the influence of interventions upon the fire resistance.

*CausalNex*
The majority of existing causal methods are often limited in their ability to consider the complex relationships between variables, i.e., assume constant/fixed relationship effects. CausalNex is a new Python library that addresses this challenge by allowing data scientists to develop models that consider how *changes* in one variable may affect *other variables* using Bayesian Networks [48]. This not only allows us to find realistic relationships between variables but also documents such relationships via causal models. A causal model that depicts the relationship between variables is displayed via directed acyclic graphs (DAGs).

*DoWhy*
The DoWhy [46] causal inference library provides an intuitive solution to estimate the average causal effect of one variable on another or upon the outcome of interest. This package offers a principled end-to-end library that can efficiently conduct causal analysis through an integrated approach that combines already existing causal and statistical methods.

**Causal Case Study**
This section describes our case study, as well as findings from the presented causal analysis. We start with a description of the adopted database and then dive into discussing our research.

*Description of database*
The adopted database homes data from 144 fire-exposed RC columns. All columns were tested in full scale and under standard fire conditions, as noted in their respective publications [14,21,42,50–60]. In this database, the following variables were collected for each RC column 1) column width, *W*, 2) steel reinforcement ratio, *r*, 3) column length, *L*, 4) concrete compressive strength, $f_c$, 5) column effective length factor, *K*, 6) concrete cover to steel reinforcement, *C*, 7) the magnitude of applied loading, *P*, and 8) fire resistance time, *FR*.

The reader is to note that this particular database was heavily used in previous ML-based papers by the authors and hence presents an attractive solution to compare the results of the presented causal analysis against that of blackbox ML analysis [42], as well as explainable ML analysis [39], and unsupervised learning analysis[3]. Further statistical details can be seen in Fig. 5 and Table 1. A look into the distribution of the variables, as shown in the histograms and provided table, shows that the selected columns are of practical ranges applicable to buildings.

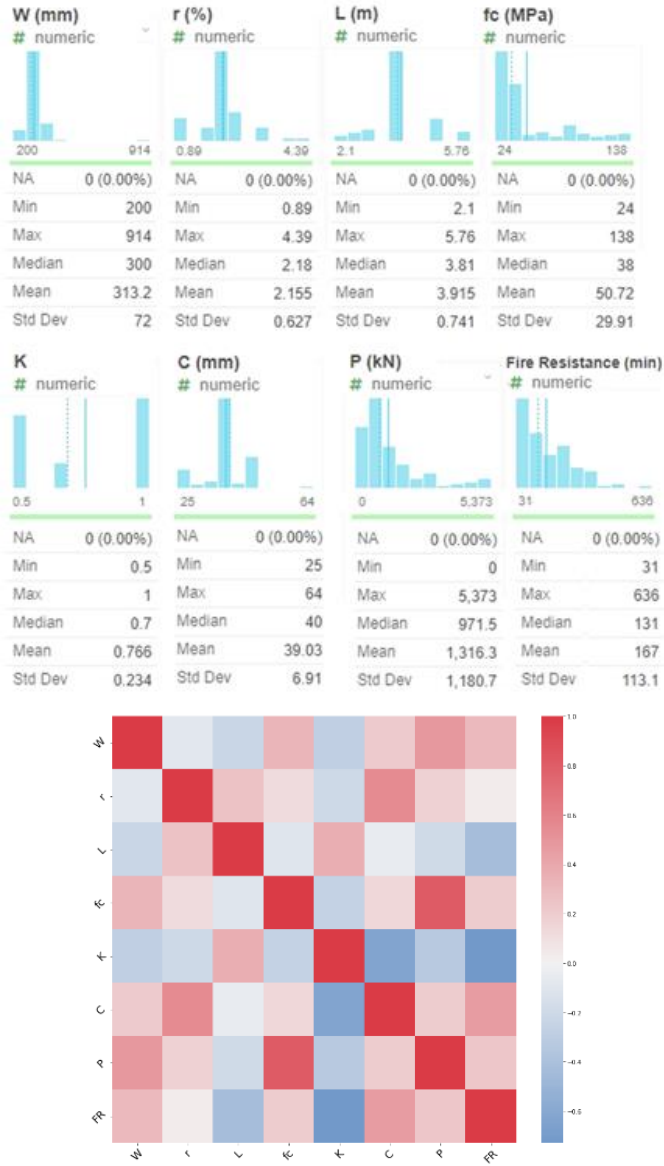

Fig. 5. Statistical insights to the compiled database (top: frequency, bottom: correlation matrix)

---

[3] This particular paper is currently under review.

Table 1 Statistics on collected database

|  | $W$ (mm) | $r$ (%) | $L$ (m) | $f_c$ (MPa) | $C$ (mm) | $P$ (kN) | $FR$ (min) |
|---|---|---|---|---|---|---|---|
| Minimum | 203 | 0.9 | 2.1 | 24 | 25 | 0 | 55 |
| Maximum | 610 | 4.4 | 5.7 | 138 | 64 | 5373 | 389 |
| Average | 350.4 | 2.1 | 3.9 | 55.7 | 42.4 | 1501.8 | 176.6 |
| Standard Deviation | 105.3 | 0.5 | 0.5 | 33 | 7.1 | 1168.6 | 82 |
| Skewness | 1.1 | 1 | -0.5 | 0.9 | -1 | 1.3 | 0.4 |

*Causal discovery analysis*

As mentioned above, we start our analysis by using Causalnex to identify the possible relationships between all collected variables and their contribution toward improving or reducing a given RC column's resistance to fire. Following the classification procedure, the discretized data is subjected to regression in order for its explanatory power. The discretized dataset has a 30% test and an 70% training split. The Random Forest machine learning model is able to achieve an $R^2$ score of 0.85 and 0.811 on the training and test data, which demonstrates its ability and potential for use in future applications where accuracy matters most.

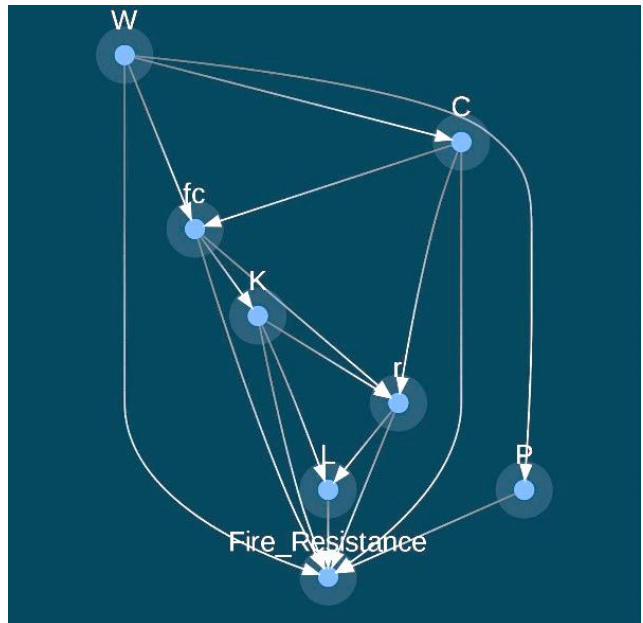

Fig. 6 Causal inference model for Fire resistance RC columns

*Causal inference analysis*

Once the causal structure was arrived at from CausalNex, we turned gears toward DoWhy to infer the outcome of possible interventions upon the fire resistance of RC columns. Four sub-analyses were conducted herein. These analyses include interventions on: isolated DAGs, DAG from CausalNex, modifying CausalNex's DAG with domain knowledge, and a hypothetical DAG. The results from each sub-analysis are presented below.

Isolated DAGs

In the first sub-analysis, we built DAGs that only tie a particular variable to the outcome ($FR = Y$) via an intervention or treatment ($T$). Fundamentally, a chain is created to link an input variable to the FR. The goal of this analysis is to explore the maximal influence of a particular variable on the fire resistance of RC columns if an intervention is applied ($T=1$), or not ($T=0$) – see Fig. 7. The value of the selected treatment for all of these models is based on the dataset's average value of the influencing variable. For example, in Fig. 7, the Treatment value is 1 if $W$ is more than 313.2 mm and is 0 for columns with $W$ is less than 313.2 mm.

For example, the model assigns $T = 1$ for all columns with $W > 313.2$ mm and also assigns $T = 0$ for all columns with $W < 313.2$ mm (while keeping the other variables as is). Now, the difference in the estimated FR for columns $T = 1$ and $T = 0$ is the tabulated mean value listed in Table 2. To ensure that this value is significant, the p-value associated with each estimated mean is checked against the traditional significance limit of 5%. Estimates with a p-value less than 5% imply significance, and p-values of more than 5% imply poor significance. In the latter, the estimated mean can be ignored.

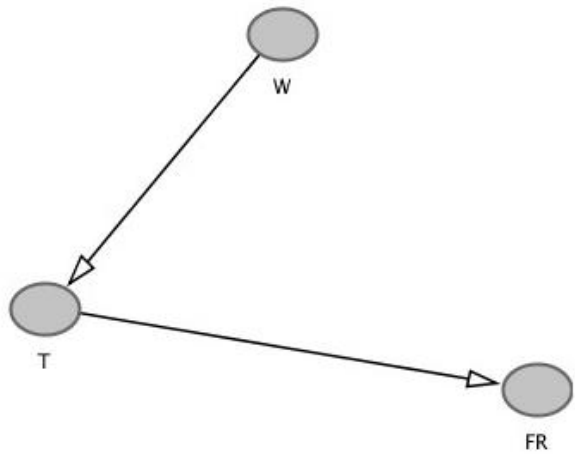

Fig. 7 Sample models of isolated variables [Note: $T$: intervention/treatment, $FR$: fire resistance]

Table 3 provides a complete picture of the carried out analysis herein. For instance, when the influence of $W$ on $FR$ is examined, the causal analysis shows that $FR$ increases by 52.8 min when $W$ is larger than the average. Similarly, when the intervention ($T = 1$) is applied, FR increases by 34.7, 55.4, 145.8, 94.4, 130.9, and 88.7 min, respectively, for $r$, $f_c$, $C$, $P$, $L$, and $K$, respectively.

Table 3 Results of analysis for fire resistance (min) [bolded p-values imply significance]

| Treatment variable | Estimate | | Refute | | |
|---|---|---|---|---|---|
| | Mean value | p-value | Random Common Cause | Data Subset Refuter | Placebo Treatment |
| $W$ | 52.8 | **0.04** | 52.6 | 52.7 | -0.48 |
| $r$ | 34.7 | 0.07* | 34.8 | 34.4 | 1.02 |
| $L$ | -130.9 | **1.4e-09** | -131.0 | -130.9 | -3.92 |
| $f_c$ | 55.4 | **0.01** | 55.4 | 55.6 | -1.15 |
| $K$ | -88.7 | **3.1e-25** | -88.7 | -88.9 | -0.23 |
| $C$ | 145.8 | **3.7e-18** | 145.6 | 146.2 | -3.48 |
| $P$ | 94.4 | **1.7e-07** | 94.2 | 94.2 | -2.20 |

* This value is close to 0.05.

The results of the analysis were examined through three different refute methods as described in the previous section. The predicted fire resistance values from the Random common cause and Data subset refuter models from three separate refute models are close to the estimated values, whereas the new values obtained from the Placebo treatment refutation model are near to zero. Thus, Table 3 shows that predictions from our analysis for each variable are reliable.

DAG from CausalNex

Second, we combine both DoWhy and CausalNEx. We employ the CausalNex-developed DAG causality model shown in Figure 8 in the DoWhy analysis. For this, we build seven distinct models, as shown in Figure 8, to examine the effect of intervention/treatment ($T$) on each variable and fire resistance. The same figure demonstrates that all of the variables, together with the treatment values, have an effect on the output (FR), and some of the variables influence each other. It is clear that some of the identified relationships do not align with our domain knowledge. Thus, for the sake of discussion, we explore the holistic causal DAG here and then refine our DAG in the next section to incorporate domain knowledge.

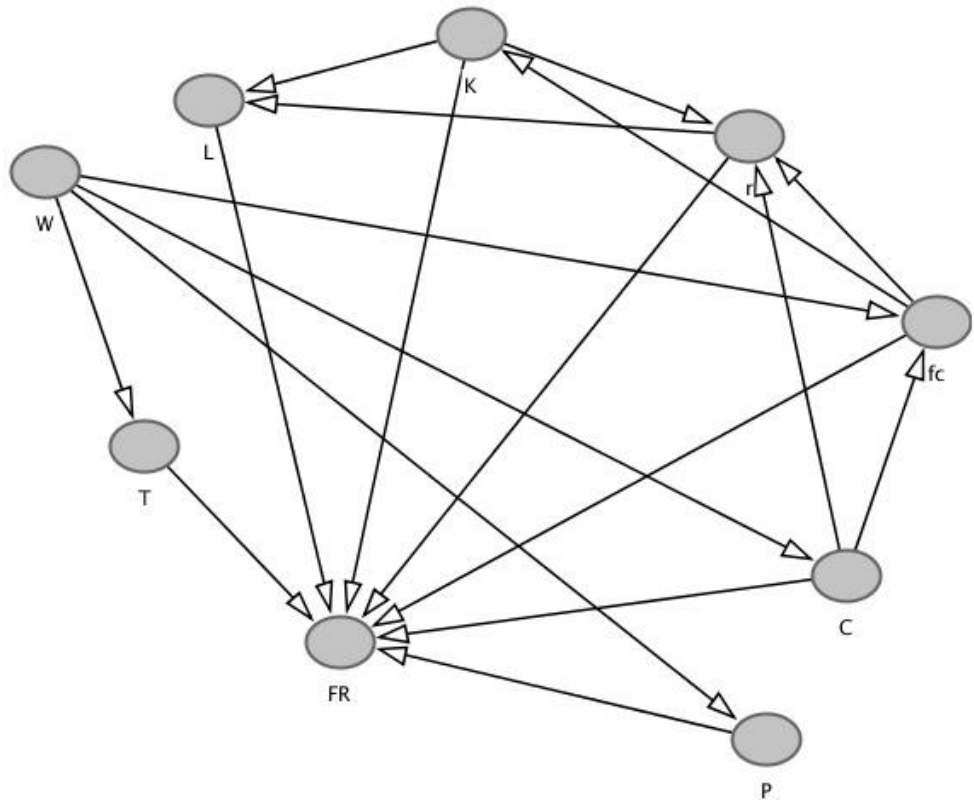

Fig. 8 CausalNex model [Note: $T$: intervention/treatment, *FR*: fire resistance]

Table 4 shows the value of fire resistance estimates and refutes for each variable. One can see that the FR reduces by 39.6 min when $T = 1$ for the steel reinforcement ratio variable. Furthermore, when all variables' effects on each other and FR are considered, it is observed that a treatment has a negative impact on *FR* for *L*, $f_c$, and *K* and a positive one on the other variables. In addition, the refute analysis confirms the reliability of these results. It is worth noting that while the mean value for W is displayed as -252 min, this value is associated with a large p-value and hence can be ignored.

Table 4 Results of analysis for fire resistance (min) [bolded p-values imply significance]

| Treatment variable | Estimate | | Refute | | |
|---|---|---|---|---|---|
| | Mean value | p-value | Random Common Cause | Data Subset Refuter | Placebo Treatment |
| W | -252.0 | 0.260 | -249.8 | -275.3 | -1.8 |
| r | 39.6 | **0.001** | 39.61 | 39.7 | 1.4 |
| L | -97.6 | **0.019** | -97.4 | -96.8 | 0.6 |
| $f_c$ | -100.7 | 0.186 | -100.8 | -101.8 | 1.2 |
| K | -60.3 | 0.688 | -60.5 | -49.3 | 1.5 |
| C | 89.4 | **3.9e-08** | 89.0 | 90.2 | 0.7 |
| P | 38.2 | **0.0001** | 38.4 | 37.7 | 2.1 |

Modifying CausalNex's DAG with domain knowledge

Given that some of the identified causal links in the DAG by CausalNex do not align with our domain knowledge, an attempt was carried out herein to augment this DAG. Thus, new DAGs were developed, as shown in Fig. 9. In creating these DAGs, we assumed the following:

- Knowing the boundary conditions of a particular column affects its length via $K$.
- Knowing the applied level of loading, $P$, influences the design of columns in terms of $f_c$, $W$, and $r$. In turns, $f_c$ also affects $r$.
- Knowing $W$, influences the size of the concrete cover, $C$.
- Given the above, links can be tied to the fire resistance ($FR = Y$).

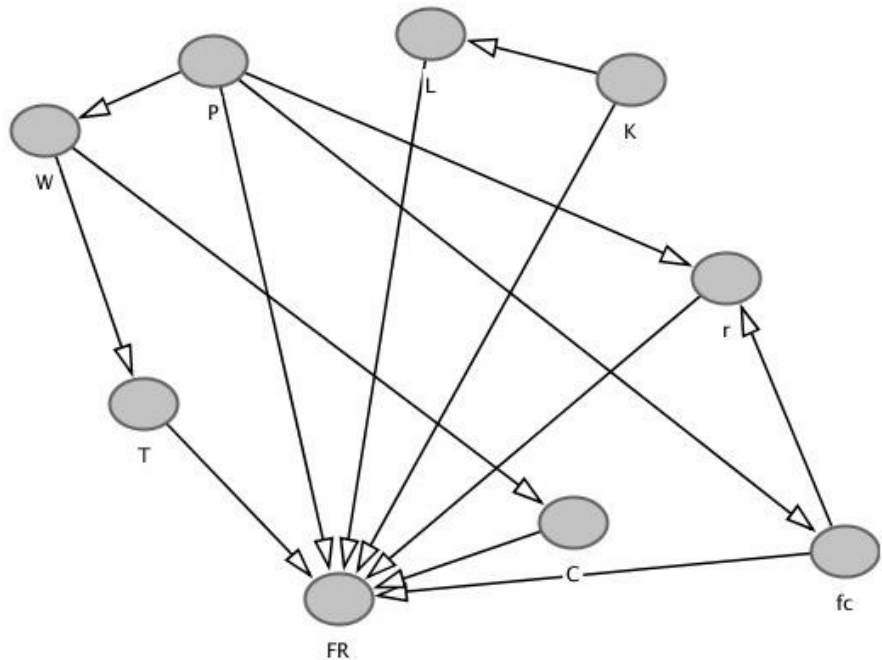

Fig. 9 Domain-knowledge revised CausalNex model [Note: $T$: intervention/treatment, $FR$: fire resistance]

As shown in Table 5, when the effects of all variables on each other and on *FR* are considered together with the effect of $T = 1$ on *FR*, then *W*, *L*, $f_c$, and *K* has a negative effect on *FR*, whereas the opposite is true for *r*, *C*, and *P*. Likewise, when the refute values are evaluated, it is clear that the estimated values are trustworthy.

Table 5 Results of analysis for fire resistance (min) [bolded p-values imply significance]

| Treatment | Estimate | | Refute | | |
|---|---|---|---|---|---|
| variable | Mean value | *p*-value | Random Common Cause | Data Subset Refuter | Placebo Treatment |
| *W* | -276.2 | 0.18 | -276.6 | -282.4 | -1.88 |
| *r* | 25.0 | 0.78 | 24.8 | 24.2 | -2.12 |
| *L* | -86.9 | 0.78 | -86.9 | -86.5 | 2.36 |
| $f_c$ | -73.1 | 0.54 | -73.3 | -79.0 | -4.71 |
| *K* | -69.6 | **1.9e-05** | -69.6 | -69.6 | 0.003 |
| *C* | 79.9 | **5.2e-09** | 79.7 | 80.4 | -0.13 |
| *P* | 44.4 | **0.01** | 44.1 | 43.6 | -2.69 |

Hypothetical DAG

Finally, we obtain seven different models by disregarding the effects of all variables on each other and assuming that they only have an influence on *FR* and that each variable above or below the mean separately affects the Treatment value, as shown in Fig. 10. In this DAG, we assumed that all variables only have a direct causal link with *FR* (i.e., without any inter-relation to other variables). Our goal is to further examine the validity of the previous DAG.

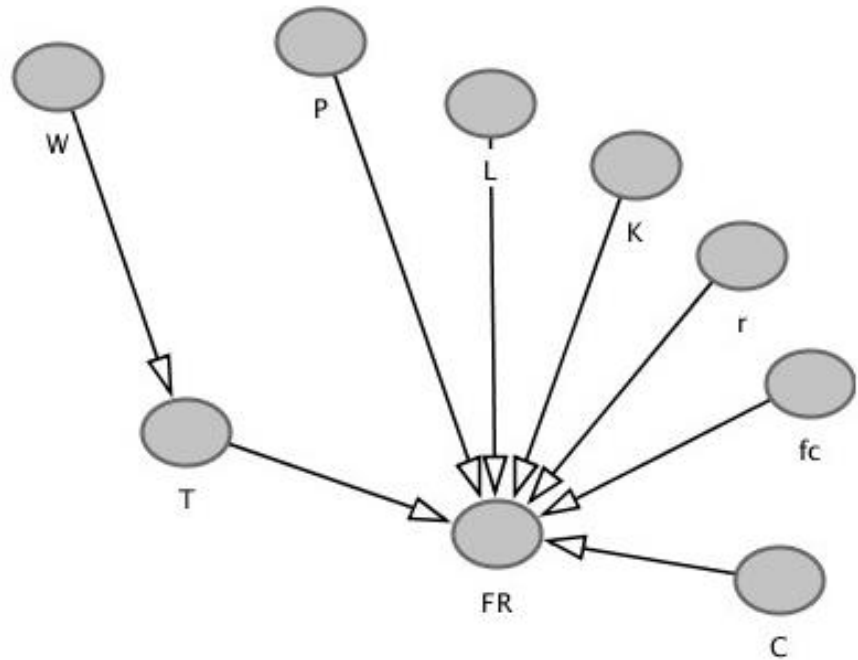

Fig. 10 Hypothetical model [Note: *T*: intervention/treatment, *FR*: fire resistance]

Table 6 shows that when all variables are assessed for their impacts on FR, positive interventions/treatments negatively influence FR for *W*, *L*, and *K*, whereas they positively influence FR for *r*, $f_c$, *C*, and *P*. The analysis from this DAG also seems to satisfy all refuting models. It is interesting to note that results from this sub-analysis match well (with the exception of $f_c$) with that from the DAG that was augmented with domain knowledge.

Table 6 Results of analysis for fire resistance (min) [bolded p-values imply significance]

| Treatment variable | Estimate | | Refute | | |
|---|---|---|---|---|---|
| | Mean value | *p*-value | Random Common Cause | Data Subset Refuter | Placebo Treatment |
| W | -245.0 | 0.11 | -247.3 | -225.7 | -2.1 |
| r | 19.0 | 0.76 | 19.2 | 18.4 | 0.27 |
| L | -82.0 | 0.97 | -81.7 | -78.0 | -2.1 |
| $f_c$ | 40.9 | 0.85 | 42.2 | 41.3 | 0.28 |
| K | -81.1 | **0.02** | -81.1 | -80.5 | 1.73 |
| C | 87.8 | **5.8e-9** | 87.9 | 87.6 | -0.07 |
| P | 36.3 | **0.004** | 36.4 | 36.3 | 0.99 |

**Comparison between Regression, Accepted Methods, Machine Learning, and Causal Analysis**

In the section, a comparison is drawn between predictions obtained from the four causal models, as well as linear regression, accepted methods, and a previously published machine learning model [39]. We start by showcasing validation plots of fire resistance predictions from each method against those measured from fire tests. These validation plots will show the predictive capability of each method. Then, all methods will be compared in terms of interventions as a means to draw attention to the key differences between predictive modeling and causal modeling.

*Predictive validation*

The multi-linear regression returned the following expression (in minutes):

$$RF = 0.35 f_c + 0.15W - 15.8r - 289.3K - 25.3L + 1.94C - 0.01P + 396.1 \qquad \text{Eq. 4}$$

It is worth noting that this expression yielded a coefficient of determination = 0.58 and a correlation coefficient of 0.76.

A full discussion on Eurocode 2 (EC2) method [62], Kodur and Raut (K&R) method [63], and machine learning (ML) [39] can be found in their respective resources. These were not repeated herein for brevity. Figure 11 depicts a comparison between all these methods. This figure clearly shows the superior predictivity of the ML model. It is also clear that all other methods, with the expectation of the Regression method, seem to perform well for columns with fire resistance of less than 240 min.

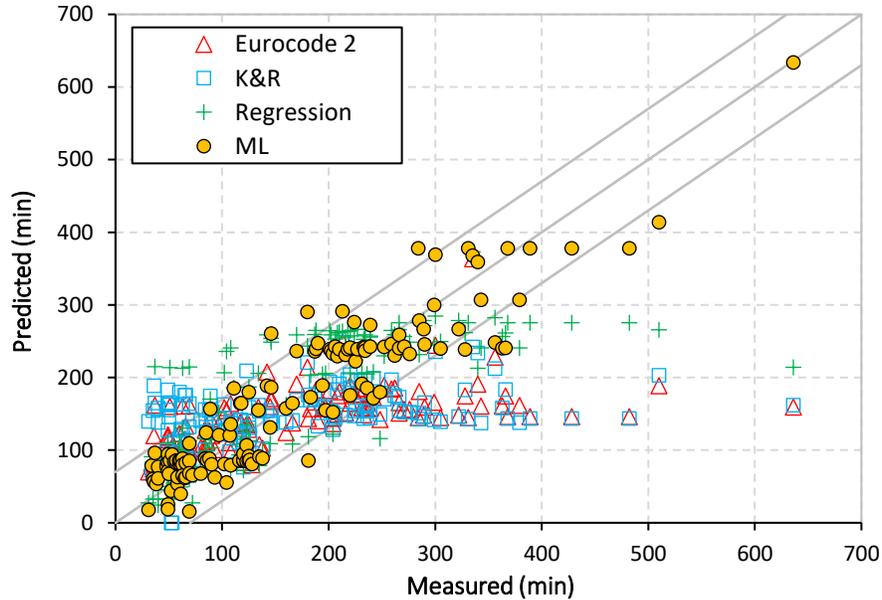

Fig. 11 Comparision between fire resistance prediction methods

*Interventional estimation*

This section draws a comparison as to how predictions from each would be affected by an act of intervention similar to that carried out in the causal case study (i.e., wherein the average value of a variable replaces the values of this particular variable across all examined columns).

Figure 12 shows how intervening was not properly captured across the methods. In fact, this action can be seen to cause a radical shift in each of the method's predictions. Such a shift can be clearly seen in the case of the most precious ML model[4]. Such a shift can be explained by the fact that all methods were designed to relate all variables, as observed in fire tests. In such tests, the columns had different variables (vs. a fixed variable across all columns). Thus, each method is primarily driven by association and correlation to minimize the variance of the outcome instead of displaying the actual causal mechanism tying each variable to the fire resistance of RC columns. This further emphasized the need to pursue causal analysis in our area.

---

[4] Please note that Fig. 12 shows a shift associated with intervening on one variable. A multi-intervention can yield a more compound shift where each method seems to center out – see sub-figures g and e.

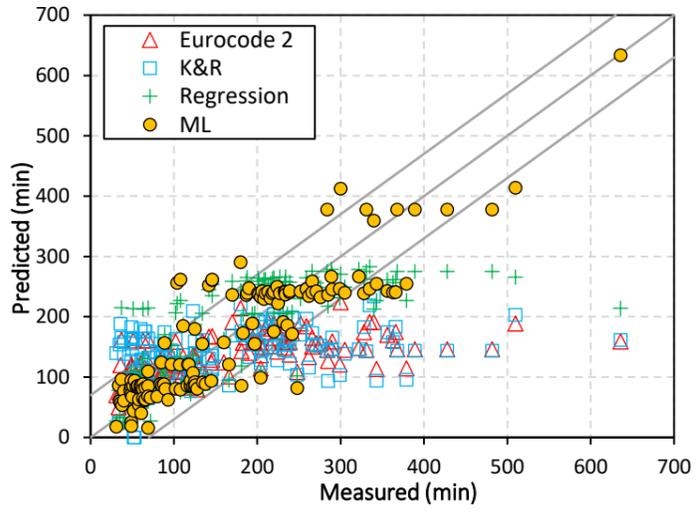

(a) $W$

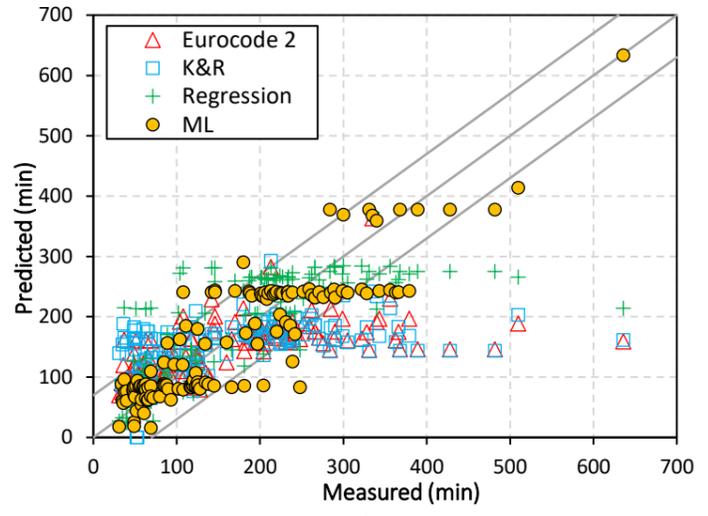

(b) $P$

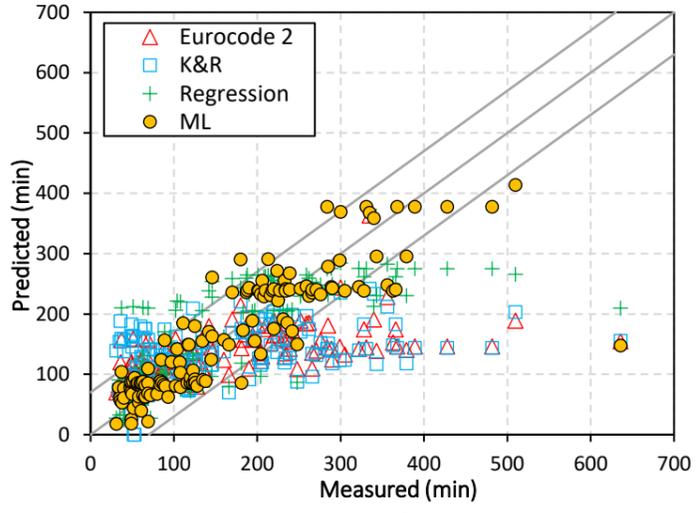

(c) $f_c$

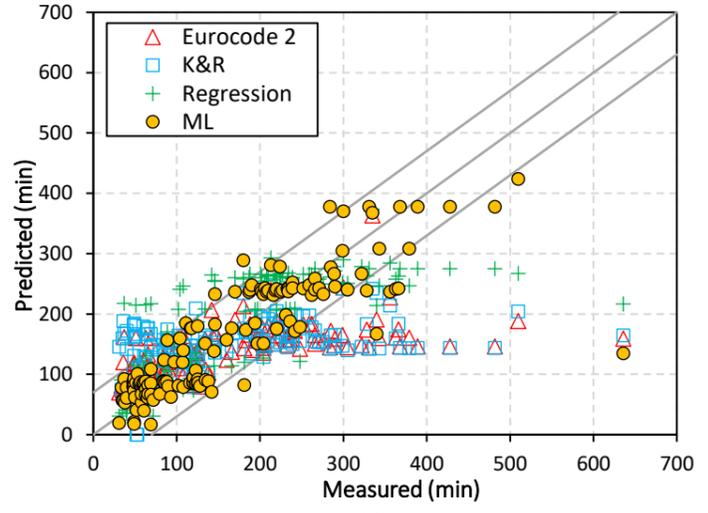

(d) $r$

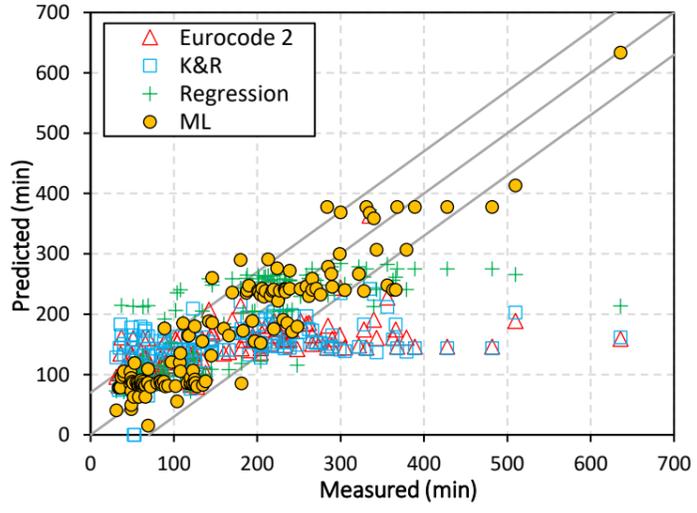

(e) $L$

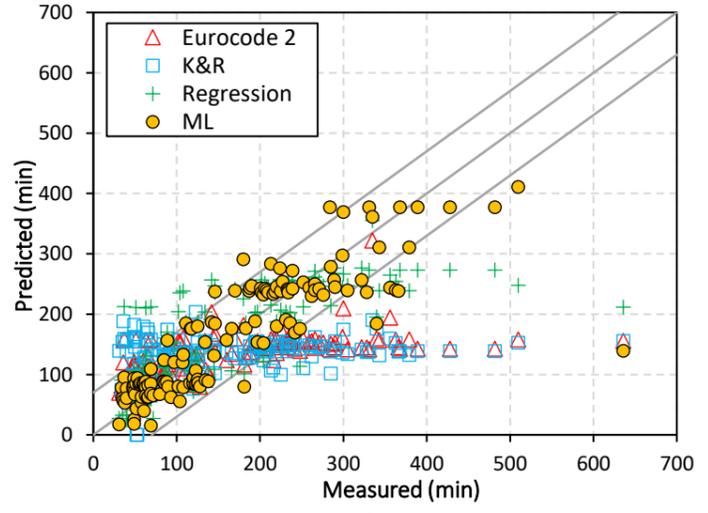

(f) $C$

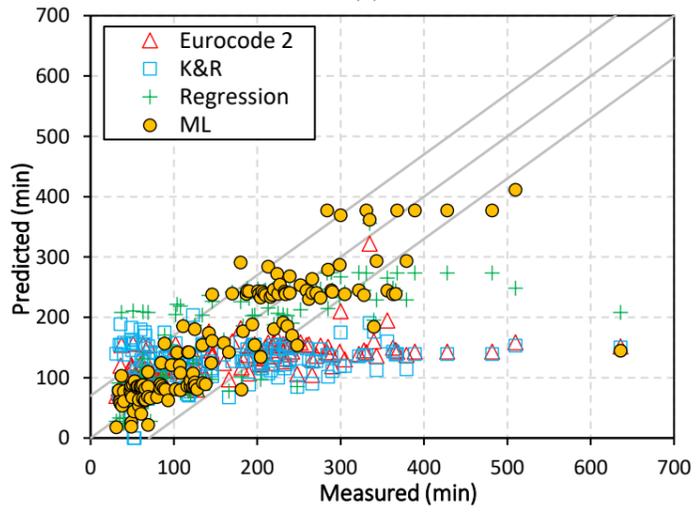

(g) $C$ and $f_c$

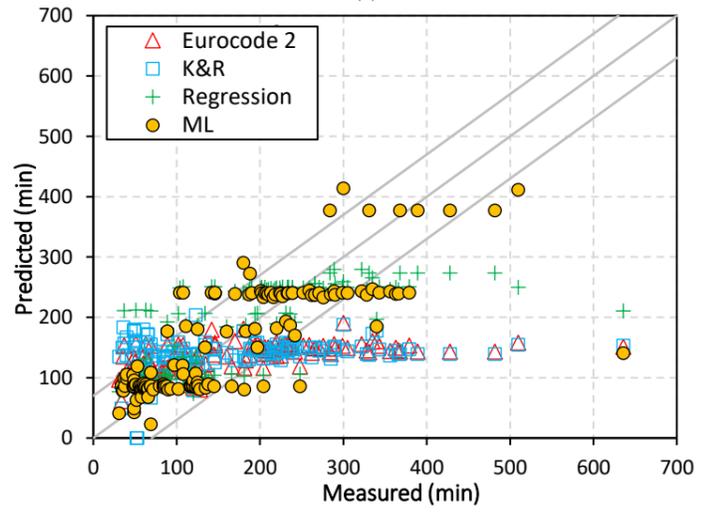

(h) *All variables*

Fig. 12 Comparisions applied due to interventions

**Conclusions**

This paper presents a look into causal discovery and causal inference as means to explore the phenomenon of fire resistance of reinforced concrete (RC) columns. Causal discovery is adopted first to uncover the causal structure regarding the examined phenomenon, and then the causal inference is applied to infer the influence of interventions on each of the identified parameters of influence. The following list of inferences can also be drawn from the findings of this study:

- Integrating causal principles is expected to further accelerate knowledge discovery in our domain. Further work on this front is warranted and certainly needed.
- Algorithmic causal graphs may, and sometimes may not, agree with domain knowledge, and hence incorporating such knowledge into casual analysis is of merit.
- Interventions upon the discovered DAGs are seen to be highly influential in terms of column width, column length, concrete cover to reinforcement, effective length factor, and compressive strength of concrete. Interventions on the level of applied loading and/or reinforcement ratio did not significantly alter fire resistance.
- Unlike traditional ML analysis, causal analysis provides us with the most realistic predictions as it can accommodate interventions (without needing new tests or experiments) vs. pure statistical associational predictions provided by traditional ML.

**Data Availability**

Some or all data, models, or code that support the findings of this study are available from the corresponding author upon reasonable request.

**Acknowledgment**

We would like to thank the Editor and Reviewers for their support of this work and for constructive comments that enhanced the quality of this manuscript.

**Conflict of Interest**

The authors declare no conflict of interest.

Version 1.0 [April 2022]